\documentclass{article}
\usepackage{spconf,amsmath,graphicx}
\usepackage{url}
\usepackage{booktabs}
\usepackage{multirow}

\title{Efficient Universal Models for Medical Image Segmentation via Weakly Supervised In-Context Learning}
\name{
  Jiesi Hu$^{1,2}$,
  Yanwu Yang$^{3,4}$,
  Zhiyu Ye$^{2}$,
  Jinyan Zhou$^{1,2}$,
  Jianfeng Cao$^{1}$,
  Hanyang Peng$^{2}$,
  Ting Ma$^{1,2}$
}

\address{
  $^{1}$ Harbin Institute of Technology at Shenzhen, Shenzhen, China \\
  $^{2}$ Peng Cheng Laboratory, Shenzhen, China \\
  $^{3}$ University Hospital Tübingen, Tübingen, Germany \\
  $^{4}$ German Center for Mental Health, Germany
}
\begin{document}
%
\maketitle

\begin{abstract}
Universal models for medical image segmentation, such as interactive and in-context learning (ICL) models, offer strong generalization but require extensive annotations. Interactive models need repeated user prompts for each image, while ICL relies on dense, pixel-level labels. To address this, we propose Weakly Supervised In-Context Learning (WS-ICL), a new ICL paradigm that leverages weak prompts (e.g., bounding boxes or points) instead of dense labels for context. This approach significantly reduces annotation effort by eliminating the need for fine-grained masks and repeated user prompting for all images. We evaluated the proposed WS-ICL model on three held-out benchmarks. Experimental results demonstrate that WS-ICL achieves performance comparable to regular ICL models at a significantly lower annotation cost. In addition, WS-ICL is highly competitive even under the interactive paradigm. These findings establish WS-ICL as a promising step toward more efficient and unified universal models for medical image segmentation. Our code and model are publicly available at https://github.com/jiesihu/Weak-ICL.
\end{abstract}

\section{Introduction}
\label{sec:intro}

\begin{figure}[t]
\centering
\includegraphics[width=0.472\textwidth]{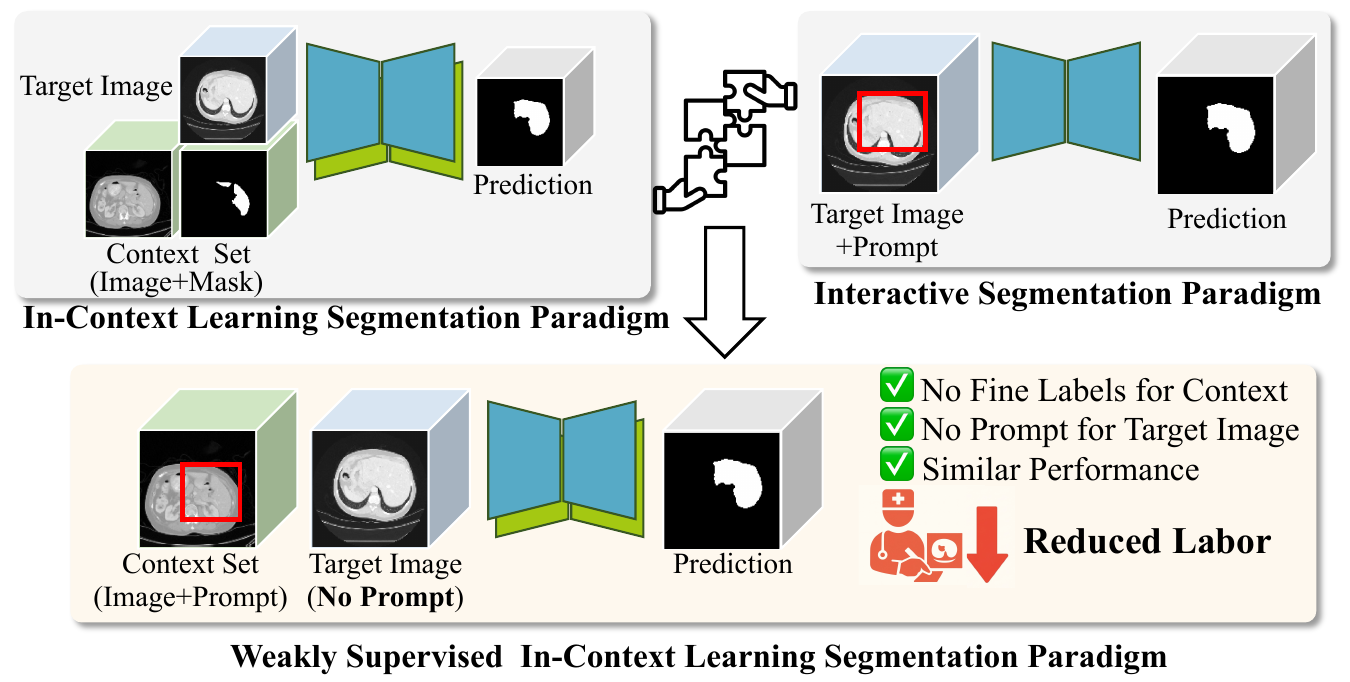}
\caption{
Comparison of segmentation paradigms. Regular in-context learning segmentation requires fine-grained masks for the context set, while interactive segmentation relies on per-image prompts. Our proposed weakly supervised ICL paradigm integrates the strengths of both approaches by using prompts in the context set, eliminating the need for fine-grained annotations or repeated prompting.
}
\label{fig:intro_wsicl}
\end{figure}

Medical image segmentation, a cornerstone of biomedical research and clinical practice, has seen a significant advancement with the emergence of universal models~\cite{butoi2023universeg, ma2024segment, hu2025medverse}. These models generalize across diverse modalities, anatomies, and clinical centers without task-specific fine-tuning, paving the way for practical AI in medicine~\cite{hu2024chebyshev}. Within this domain, two paradigms have become prominent: interactive models~\cite{ma2024segment,isensee2025nninteractive, yang2025medsamix} and in-context learning (ICL) models~\cite{butoi2023universeg, hu2025building, hu2024icl}. While both are powerful, they are tailored for distinct application scenarios.

Interactive models leverage simple user prompts, such as points or bounding boxes, as weak supervision to generate precise segmentation masks. This paradigm was successfully applied to 2D medical imaging by the SAM family~\cite{kirillov2023segment, ma2024segment}. Recent work has extended these models to 3D~\cite{wang2024sam}, with frameworks like nnInteractive further boosting 3D performance and supporting more diverse interaction types~\cite{isensee2025nninteractive}. Despite these advances, a core limitation persists: a new prompt is required for every individual image, making the process inefficient for segmenting large datasets.

ICL models perform segmentation by referencing a context set, which consists of a few example images paired with fine-grained masks to guide the processing of unlabeled images. Unlike interactive models, ICL avoids the need to provide prompts for every individual image. This paradigm, first demonstrated on natural images with models like SegGPT~\cite{wang2023seggpt}, was subsequently adapted for 2D medical imaging~\cite{butoi2023universeg, czolbe2023neuralizer} and recently extended to 3D with frameworks such as Neuroverse3D~\cite{hu2025building}. Unlike interactive methods, ICL requires no per-image manual effort once the context set is established. However, its drawback lies in the costly construction of the context set, as fine-grained annotations are needed, which is particularly burdensome.

Inspired by the success of both paradigms, we introduce \textbf{Weakly Supervised ICL Segmentation (WS-ICL)}, a new paradigm where the context set is built using only weak supervision like bounding boxes or points, rather than dense masks (Fig.~\ref{fig:intro_wsicl}). This hybrid approach captures the strengths of both prior methods. It eliminates the need for laborious, fine-grained annotations typical of ICL, while retaining the prompt-once-segment-many efficiency, avoiding the repetitive prompting required by interactive models.

To validate this concept, we build and train universal WS-ICL models on 18 medical imaging datasets and evaluate on 3 held-out datasets. Our experiments show that the models' performance is comparable to that of fully supervised ICL, but at a fraction of the annotation cost. Furthermore, our models inherently function as high-performing interactive models, achieving results near the state-of-the-art. Our contributions are:
\begin{itemize}
    \item We propose and validate WS-ICL, a new universal segmentation paradigm that leverages in-context weak supervision to drastically reduce annotation labor.
    \item We introduce a dual-branch U-Net model capable of operating in both the WS-ICL and interactive paradigms.
    \item Through comprehensive evaluations on held-out datasets, we demonstrate that our models are competitive with fully supervised ICL and approach state-of-the-art performance in the interactive paradigm.
\end{itemize}

\section{Method}
We construct a model that segments a target image $x$ conditioned on a context set $S$, where $S$ is composed of image–prompt pairs $\{(x_i, u_i)\}_{i=1}^{L}$. Here, $x_i$ denotes an input image, $u_i$ represents a weak prompt (e.g., a bounding box or a point) rather than a dense segmentation mask as in regular ICL, and $L$ is the size of the context set. In this weakly supervised in-context learning setting, we aim to learn a universal function $\hat{y} = f_{\theta}\!\left(x, S\right)$ that predicts a segmentation map $\hat{y}$ for the target image $x$, conditioned on the task-specific context set. By design, the prompts in $S$ provide coarse supervision that specifies the approximate location of the segmentation target, while the target image $x$ is processed without any prompt.

\begin{figure}[t]
\centering
\includegraphics[width=0.472\textwidth]{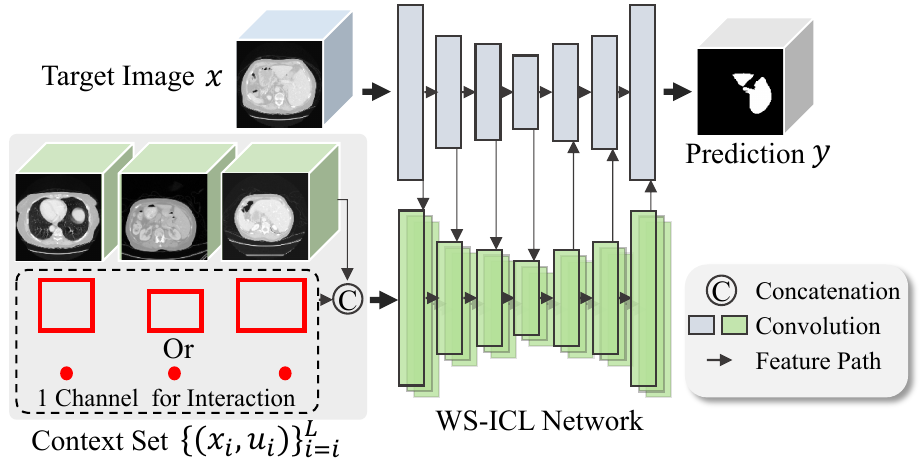}
\caption{Illustration of the proposed WS-ICL task. Context images are concatenated with the prompt channels and jointly processed with the target image through the WS-ICL network to generate the segmentation prediction.}
\label{fig:main_fig}
\end{figure}

\subsection{Model}
\vspace{0.4em}\noindent\textbf{Network Architecture.} 
Figure~2 illustrates our WS-ICL segmentation pipeline. We adopt the architecture of Neuroverse3D~\cite{hu2025building}, a state-of-the-art ICL model for 3D medical imaging, as our backbone. Its network features a dual-branch design for target and context inputs, with cross-branch feature interactions at each layer. Notably, the network is memory-efficient, enabling it to process a large number of context images with a small memory requirement. To incorporate prompts, we follow the strategy of nnInteractive~\cite{isensee2025nninteractive}, which encodes the prompt as an additional input channel. Consequently, the input to our context branch is a two-channel tensor containing both the image and its corresponding weak prompt. This design enables the model to learn task-relevant representations from the highest-resolution features.

\vspace{0.4em}\noindent\textbf{Loss Function.} 
For a fair comparison, we adopt the same modified $\text{smooth}-L_{1}$ loss used in Neuroverse3D~\cite{hu2025building} as the segmentation loss during training.

\vspace{0.4em}\noindent\textbf{Operation in Interactive Mode.} 
By feeding the same image to both the target and context branches and supplying the user prompt through the context branch, the model can function as an interactive model without any modification to its structure or weights.

\subsection{Prompting}
We train two separate weights, one for bounding box prompts and another for point prompts. Additionally, our models support multiple number of prompt for a single image to specify more details. During training, we generate both bounding box and point interactions through simulation. Inspired by the reading habits of radiologists~\cite{lee2024read}, we first sample 2D slices from the 3D volume on which to place the prompt. The sampling probability for each slice is linearly proportional to the area of the target region it contains. The specific interaction simulations are performed as follows.

\vspace{0.4em}\noindent\textbf{Bounding Box Interactions.}
Although our model directly processes 3D medical images, it accepts 2D bounding box prompts. This approach avoids including excessive non-target regions and offers a more user-friendly interaction for radiologists~\cite{isensee2025nninteractive}. For each selected 2D slice mask, we first perform connected component analysis and generate a tight bounding box for each component. To introduce variability, a rounded random variable $g \sim \mathcal{N}(0,1)$ is added to each bounding box coordinate. Finally, the bounding box is rendered as a filled rectangle into the prompt input channel.

\begin{table*}[htbp]
\centering
\renewcommand{\arraystretch}{1.0}
\setlength{\tabcolsep}{4pt}
\caption{Comparison of our proposed WS-ICL with state-of-the-art ICL models on three held-out datasets, reported in Dice coefficient (\%). Supervision in the context set specifies the type and number of annotations or prompts. M. Sinus refers to Maxillary Sinus. The context set consists of 8 3D images for all models.}
\resizebox{0.99\textwidth}{!}{
\begin{tabular}{l |c| cccc |cccc| cc| c}
\toprule
 \multirow{2}{*}{\textbf{Methods }}
&   \multirow{2}{*}{\renewcommand{\arraystretch}{0.8}\begin{tabular}[x]{@{}c@{}}\textbf{Supervision }\\\textbf{in Context Set}\end{tabular} }
 & \multicolumn{4}{c|}{\textbf{FLARE22}~\cite{ma2024unleashing} }
 & \multicolumn{4}{c|}{\textbf{Nasal}~\cite{zhang2024nasalseg} }
 & \multicolumn{2}{c|}{\textbf{Mice}~\cite{rosenhain2018preclinical} }
 & \multirow{2}{*}{\textbf{Average}} \\

 & & Liver 
 & \renewcommand{\arraystretch}{0.8}\begin{tabular}[x]{@{}c@{}}Right\\Kidney\end{tabular}
 & \renewcommand{\arraystretch}{0.8}\begin{tabular}[x]{@{}c@{}}Left\\Kidney\end{tabular} 
 & Spleen 
 & \renewcommand{\arraystretch}{0.8}\begin{tabular}[x]{@{}c@{}}Nasal\\Cavity\end{tabular}   
 & \renewcommand{\arraystretch}{0.8}\begin{tabular}[x]{@{}c@{}}Nasal\\Pharynx\end{tabular}
 & \renewcommand{\arraystretch}{0.8}\begin{tabular}[x]{@{}c@{}}Right M.\\Sinus\end{tabular}  
& \renewcommand{\arraystretch}{0.8}\begin{tabular}[x]{@{}c@{}}Left M.\\Sinus\end{tabular}  
 & Lung & Pancreas &  \\
\midrule
\addlinespace[2.5pt]
\multicolumn{13}{l}{\textit{\textbf{Task-Specific Upper Bound}}} \\
\addlinespace[2.5pt]
nnUNet & N/A & 98.53 & 96.34 & 91.93 & 92.40 & 92.29 & 95.84 & 94.97 & 96.18 & 94.25 & 85.91 & 93.86 \\
\midrule
\addlinespace[2.5pt]
\multicolumn{13}{l}{\textit{\textbf{Regular (Fully Supervised) In-Context Learning Models}}} \\
\addlinespace[2.5pt]
SegGPT~\cite{wang2023seggpt} & 8 2D Annotations & 80.25 & 73.67 & 52.70 & 67.02 & 52.49 & 44.71 & 50.38 & 50.54 & 53.73 & 47.63 & 57.31 \\

Neuralizer~\cite{czolbe2023neuralizer} & 32 2D Annotations & 70.18 & 65.39 & 60.12 & 69.73 & 61.55 & 73.14 & 75.13 & 74.31 & 70.61 & 43.79 & 66.40 \\
UniverSeg~\cite{butoi2023universeg} & 64 2D Annotations & 82.58 & 80.46 & 57.31 & 57.57 & \underline{76.67} & 73.06 & 80.06 & 81.75 & 58.95 & 41.96 & 69.04 \\
ICL-SAM~\cite{hu2024icl} & 64 2D Annotations & 82.97 & 80.82 & 58.03 & 59.07 & 74.66 & 73.59 & 80.80 & 81.47 & 60.03 & 41.21 & 69.27 \\

Neuroverse3D~\cite{hu2025building} & 8 3D Annotations & 92.50 & 75.72 & 75.83 & 79.88 & 76.01 & \underline{86.97} & 79.32 & 83.28 & \underline{85.42} & \underline{66.74} & 80.17 \\
Neuroverse3D*~\cite{hu2025building} & 8 3D Annotations & \textbf{95.19} &	89.22 &	86.14 &	\underline{88.93} &	\textbf{79.88} &	\textbf{88.08} &	\underline{88.44} &	90.08 &	\textbf{89.04} &	60.71 &	\textbf{85.57}\\
\midrule
\addlinespace[2.5pt]
\multicolumn{13}{l}{\textit{\textbf{Weakly Supervised In-Context Learning Models}}} \\
\addlinespace[2.5pt]
WS-ICL (Box) & 40 2D Boxes & \underline{94.57} &	\textbf{93.82} &	\underline{90.66} &	\textbf{89.64} &	51.62 &	84.04 &	87.42 &	\textbf{91.89} &	79.66 &	\textbf{67.38} &	\underline{83.07}\\
WS-ICL (Point) & 40 Points & 92.13 &	\underline{89.63}	 &\textbf{92.38} &	80.88 &	40.59 &	80.3 &	\textbf{88.97} &	\underline{90.93} &	65.88	 &60.55 &	78.22\\
\bottomrule
\end{tabular}}

\label{tab:icl_comparison}
\end{table*}

\vspace{0.4em}\noindent\textbf{Point Interactions.}
For point interactions, connected component analysis is also applied to the selected slice mask, and a random point is sampled within each component. To make prompts more perceptible to the model, the sampled point is expanded into a sphere and converted into a soft mask, with maximum intensity at the center defined by a normalized Euclidean distance transform~\cite{isensee2025nninteractive}.

\section{Experiments}
\subsection{Experimental Settings}
\noindent\textbf{Dataset.} 
To ensure strong generalization, we train our models on a diverse compilation of \textbf{18} publicly available datasets, totaling \textbf{39,213} 3D scans. This collection includes all training data from~\cite{hu2025building} as well as several other datasets~\cite{ji2022amos,luo2024rethinking,wasserthal2023totalsegmentator}. The datasets cover widely used imaging modalities such as CT, T1, T2, FLAIR, MRA, DWI, ADC, and PD, and common anatomical regions such as the brain, abdomen, prostate, and lung. To further enhance generalization, we also incorporate \textbf{20,000} synthetic 3D images and corresponding masks generated using approaches introduced in~\cite{butoi2023universeg,hu2025towards}.



We assess our models' generalization to unseen distributions using three held-out datasets that pose distinct challenges, including abdominal scans from an unseen medical center (FLARE22~\cite{ma2024unleashing}, 50 3D images), segmentation of an unseen anatomical structure (Nasal~\cite{zhang2024nasalseg}, 130 3D images), and scans of an unseen species (Mice~\cite{rosenhain2018preclinical}, 40 3D images). 

\noindent\textbf{Training and Evaluation Protocol.} 
All inputs are resized to $128 \times 128 \times 128$ and normalized to the range $[0,1]$. The models are initialized with Neuroverse3D pretrained weights~\cite{hu2025building} and fine-tuned for four days on a single NVIDIA A100 80GB GPU with a learning rate of $1 \times 10^{-6}$ using the Adam optimizer. Other training protocols, such as data augmentation and task augmentation techniques for training ICL models, follow~\cite{hu2025building}.

For evaluation, model performance is assessed using the commonly adopted Dice coefficient. Each task is tested eight times for reliable evaluation, with a different context set randomly sampled for each run.

\noindent\textbf{Compared Models.} 
We compare our method with several state-of-the-art ICL models, including SegGPT~\cite{wang2023seggpt}, UniverSeg~\cite{butoi2023universeg}, Neuralizer~\cite{czolbe2023neuralizer}, and ICL-SAM~\cite{hu2024icl}, all of which are designed for 2D inputs, as well as Neuroverse3D~\cite{hu2025building}, which directly handles 3D images. The hyperparameters of these methods follow the optimal configurations reported in their respective papers. All models are evaluated using their publicly released pretrained weights. In addition, for a fair comparison, we fine-tune Neuroverse3D on our dataset, denoted as Neuroverse3D*. nnU-Net~\cite{isensee2021nnu} is trained directly on the held-out sets to serve as an upper bound.

\subsection{Results}
\noindent\textbf{Comparison with Other ICL Models.} 
As shown in Table~\ref{tab:icl_comparison}, our WS-ICL requires substantially less supervision in context compared to other ICL models, yet achieves strong performance on well-defined targets such as the liver and kidney, approaching Neuroverse3D*. This demonstrates that for many organs, effective in-context learning can be achieved without fine-grained segmentation, leading to significant savings in annotation effort. However, for targets with ambiguous boundaries, such as the nasal cavity, bounding-box- and point-based context are insufficient to convey precise segmentation intent, resulting in inferior performance. Therefore, WS-ICL should be used as an efficient first-pass approach, with the more labor-intensive regular ICL reserved for challenging targets where WS-ICL may fall short.


\begin{figure}[ht]
\centering
\includegraphics[width=0.475\textwidth]{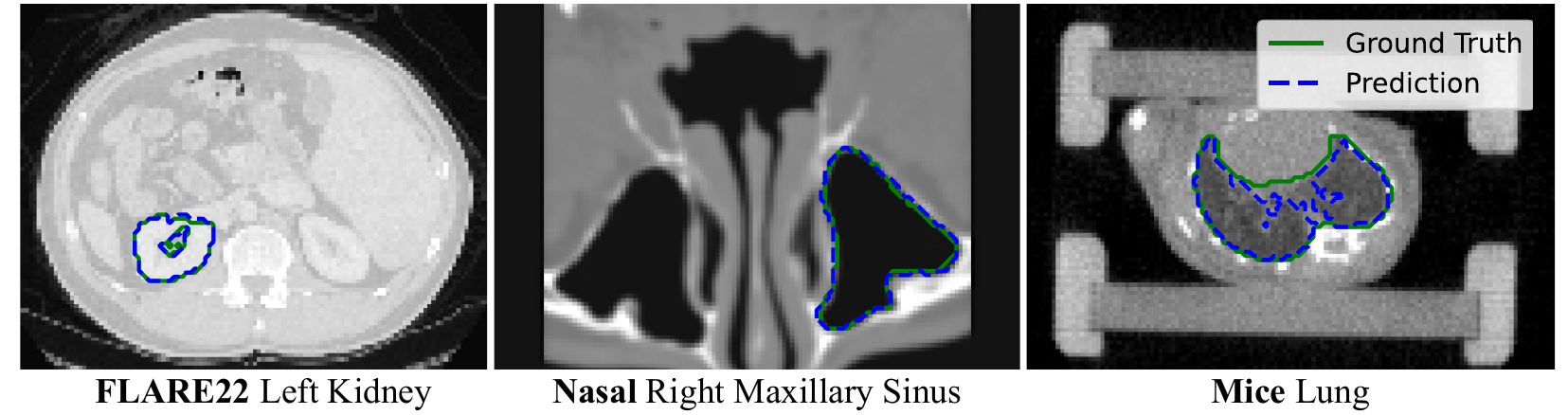}
\caption{Qualitative results of WS-ICL (Box) with 8 context images and 5 prompts per image.}
\label{fig:Qualitative}
\end{figure}

\noindent\textbf{Qualitative Results.} 
Figure~\ref{fig:Qualitative} shows that the model achieves accurate segmentation on many targets with clear boundaries, even when the context provides only coarse prompts. This further demonstrates the effectiveness of WS-ICL in diverse scenarios.

\begin{figure}[ht]
\centering
\includegraphics[width=0.472\textwidth]{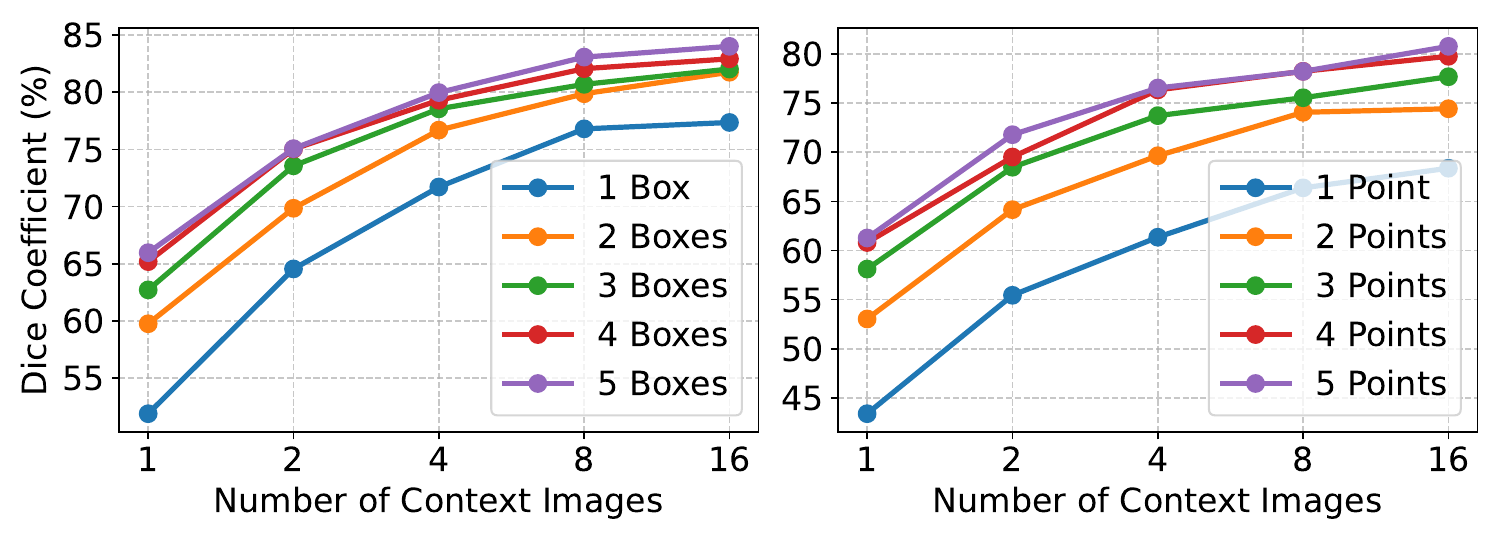}
\caption{Model performance with different context set sizes. The legend indicates the number of prompts per image. Dice scores are averaged over all tasks.}
\label{fig:context_size}
\end{figure}

\noindent\textbf{Analysis of Different Context Sizes.} 
Figure~\ref{fig:context_size} shows model performance under varying context set sizes and different numbers of prompts per image. The results follow the general trend of ICL models, where performance improves as the context size increases. A clear trend of diminishing returns is also evident, with performance gains beginning to plateau after 8 context images. Moreover, increasing the number of prompts per image from one to two yields a significant gain, highlighting the importance of providing multiple prompts for each context image. 

\begin{figure}[ht]
\centering
\includegraphics[width=0.472\textwidth]{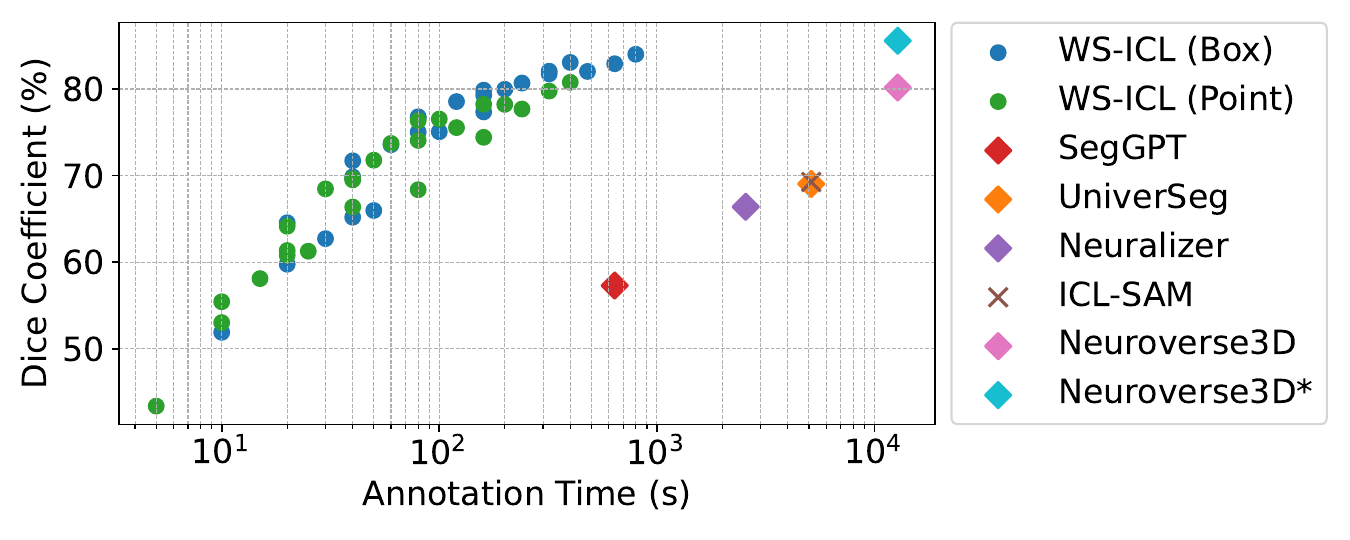}
\caption{Performance of different models and the corresponding annotation time for context construction. Annotation times are approximated as 5, 10, 80, and 1600 seconds for point, bounding box, 2D mask, and 3D mask, respectively. Dice scores are averaged over all tasks.}
\label{fig:time}
\end{figure}

\noindent\textbf{Model Efficiency Analysis.} 
Figure~\ref{fig:time} illustrates the trade-off between annotation time and model performance under various WS-ICL settings, including different context set sizes and numbers of prompts per image. We conservatively estimate the time required for 2D and 3D annotations. Even under this assumption, the results show that WS-ICL is highly efficient, achieving performance comparable to Neuroverse3D* with less than one-tenth of the annotation time. Compared to other models, it also consistently requires substantially less labor to reach similar performance levels. Furthermore, the plot shows that bounding box and point prompts cluster in the same region, suggesting they offer a similar balance between performance and annotation cost.

\begin{table}[ht]
\centering
\caption{Performance comparison of WS-ICL models in the interactive paradigm with state-of-the-art interactive models. Dice scores are averaged over tasks within each dataset. Numbers with * indicate models that have been trained on the corresponding dataset.}
\resizebox{0.475\textwidth}{!}{
\begin{tabular}{l|c|ccc}
\toprule
\textbf{Methods} & \textbf{Prompt} & \textbf{FLARE22}~\cite{ma2024unleashing} & \textbf{Nasal} ~\cite{zhang2024nasalseg} & \textbf{Mice}~\cite{rosenhain2018preclinical}  \\
\midrule
MedSAM~\cite{ma2024segment}            & Many Boxes & 85.21* & 67.98 & 63.01 \\
nnInteractive~\cite{isensee2025nninteractive}     & 1 Box      & \textbf{96.36}* & \textbf{72.91} & \textbf{77.93} \\
WS-ICL (Box)      & 1 Box      & \underline{92.03} & \underline{71.90} & \underline{74.67} \\
\midrule
SAM-Med3D~\cite{wang2024sam}          & 1 Point    & 87.30* & 32.56 & 14.59 \\
nnInteractive~\cite{isensee2025nninteractive}     & 1 Point    & \underline{90.08}* & \textbf{58.37} & \textbf{73.53} \\
WS-ICL (Point)    & 1 Point    & \textbf{90.44} & \underline{54.99} & \underline{62.52} \\
\bottomrule
\end{tabular}}

\label{tab:interactive}
\end{table}

\noindent\textbf{Performance under the Interactive Paradigm.} 
Table~\ref{tab:interactive} demonstrates that our models also exhibits strong competitiveness in the interactive paradigm, achieving performance close to nnInteractive~\cite{isensee2025nninteractive}, the current leading interactive model for medical image segmentation, and significantly outperforming MedSAM~\cite{ma2024segment} and SAM-Med3D~\cite{wang2024sam}. Crucially, this performance is achieved in a zero-shot setting, whereas competing models with an * were trained on the corresponding datasets. These results suggest that our model also provides a reliable pathway to unify the WS-ICL and interactive paradigms in a simple yet decent way.

\section{Conclusion}
In this work, we introduced WS-ICL for medical image segmentation, a new paradigm that leverages weak prompts such as bounding boxes and points in the context set instead of fine-grained annotations. By integrating the strengths of in-context learning and interactive segmentation, WS-ICL significantly reduces annotation effort while maintaining competitive performance across diverse medical imaging tasks. Extensive experiments on 3 held-out datasets demonstrate that WS-ICL achieves performance close to regular ICL models and remains highly competitive in the interactive paradigm. These results highlight WS-ICL as a promising model toward efficient ICL and provide a pathway for further unifying WS-ICL and interactive paradigms.

\bibliographystyle{IEEEbib}
\bibliography{strings,refs}

\begin{thebibliography}{10}

\bibitem{butoi2023universeg}
Victor~Ion Butoi, Jose Javier~Gonzalez Ortiz, Tianyu Ma, Mert~R Sabuncu, John
  Guttag, and Adrian~V Dalca,
\newblock ``Universeg: Universal medical image segmentation,''
\newblock in {\em Proceedings of the IEEE/CVF International Conference on
  Computer Vision}, 2023, pp. 21438--21451.

\bibitem{ma2024segment}
Jun Ma, Yuting He, Feifei Li, Lin Han, Chenyu You, and Bo~Wang,
\newblock ``Segment anything in medical images,''
\newblock {\em Nature Communications}, vol. 15, no. 1, pp. 654, 2024.

\bibitem{hu2025medverse}
Jiesi Hu, Jianfeng Cao, Yanwu Yang, Chenfei Ye, Yixuan Zhang, Hanyang Peng, and
  Ting Ma,
\newblock ``Medverse: A universal model for full-resolution 3d medical image
  segmentation, transformation and enhancement,''
\newblock {\em arXiv preprint arXiv:2509.09232}, 2025.

\bibitem{hu2024chebyshev}
Jiesi Hu, Yanwu Yang, Xutao Guo, Jinghua Wang, and Ting Ma,
\newblock ``A chebyshev confidence guided source-free domain adaptation
  framework for medical image segmentation,''
\newblock {\em IEEE Journal of Biomedical and Health Informatics}, 2024.

\bibitem{isensee2025nninteractive}
Fabian Isensee, Maximilian Rokuss, Lars Kr{\"a}mer, Stefan Dinkelacker, Ashis
  Ravindran, Florian Stritzke, Benjamin Hamm, Tassilo Wald, Moritz Langenberg,
  Constantin Ulrich, et~al.,
\newblock ``nninteractive: Redefining 3d promptable segmentation,''
\newblock {\em arXiv preprint arXiv:2503.08373}, 2025.

\bibitem{yang2025medsamix}
Yanwu Yang, Guinan Su, Jiesi Hu, Francesco Sammarco, Jonas Geiping, and Thomas
  Wolfers,
\newblock ``Medsamix: A training-free model merging approach for medical image
  segmentation,''
\newblock {\em arXiv preprint arXiv:2508.11032}, 2025.

\bibitem{hu2025building}
Jiesi Hu, Hanyang Peng, Yanwu Yang, Xutao Guo, Yang Shang, Pengcheng Shi,
  Chenfei Ye, and Ting Ma,
\newblock ``Building 3d in-context learning universal model in neuroimaging,''
\newblock {\em arXiv preprint arXiv:2503.02410}, 2025.

\bibitem{hu2024icl}
Jiesi Hu, Yang Shang, Yanwu Yang, Xutao Guo, Hanyang Peng, and Ting Ma,
\newblock ``Icl-sam: Synergizing in-context learning model and sam in medical
  image segmentation,''
\newblock {\em Medical Imaging with Deep Learning}, pp. 641--656, 2024.

\bibitem{kirillov2023segment}
Alexander Kirillov, Eric Mintun, Nikhila Ravi, Hanzi Mao, Chloe Rolland, Laura
  Gustafson, Tete Xiao, Spencer Whitehead, Alexander~C Berg, Wan-Yen Lo,
  et~al.,
\newblock ``Segment anything,''
\newblock in {\em Proceedings of the IEEE/CVF International Conference on
  Computer Vision}, 2023, pp. 4015--4026.

\bibitem{wang2024sam}
Haoyu Wang, Sizheng Guo, Jin Ye, Zhongying Deng, Junlong Cheng, Tianbin Li,
  Jianpin Chen, Yanzhou Su, Ziyan Huang, Yiqing Shen, et~al.,
\newblock ``Sam-med3d: towards general-purpose segmentation models for
  volumetric medical images,''
\newblock in {\em European Conference on Computer Vision}. Springer, 2024, pp.
  51--67.

\bibitem{wang2023seggpt}
Xinlong Wang, Xiaosong Zhang, Yue Cao, Wen Wang, Chunhua Shen, and Tiejun
  Huang,
\newblock ``Seggpt: Segmenting everything in context,''
\newblock {\em arXiv preprint arXiv:2304.03284}, 2023.

\bibitem{czolbe2023neuralizer}
Steffen Czolbe and Adrian~V Dalca,
\newblock ``Neuralizer: General neuroimage analysis without re-training,''
\newblock in {\em Proceedings of the IEEE/CVF Conference on Computer Vision and
  Pattern Recognition}, 2023, pp. 6217--6230.

\bibitem{lee2024read}
Changsun Lee, Sangjoon Park, Cheong-Il Shin, Woo~Hee Choi, Hyun~Jeong Park,
  Jeong~Eun Lee, and Jong~Chul Ye,
\newblock ``Read like a radiologist: Efficient vision-language model for 3d
  medical imaging interpretation,''
\newblock {\em arXiv preprint arXiv:2412.13558}, 2024.

\bibitem{ma2024unleashing}
Jun Ma, Yao Zhang, Song Gu, Cheng Ge, Shihao Mae, Adamo Young, Cheng Zhu, Xin
  Yang, Kangkang Meng, Ziyan Huang, et~al.,
\newblock ``Unleashing the strengths of unlabelled data in deep
  learning-assisted pan-cancer abdominal organ quantification: the flare22
  challenge,''
\newblock {\em The Lancet Digital Health}, vol. 6, no. 11, pp. e815--e826,
  2024.

\bibitem{zhang2024nasalseg}
Yichi Zhang, Jing Wang, Tan Pan, Quanling Jiang, Jingjie Ge, Xin Guo, Chen
  Jiang, Jie Lu, Jianning Zhang, Xueling Liu, et~al.,
\newblock ``Nasalseg: A dataset for automatic segmentation of nasal cavity and
  paranasal sinuses from 3d ct images,''
\newblock {\em Scientific Data}, vol. 11, no. 1, pp. 1329, 2024.

\bibitem{rosenhain2018preclinical}
Stefanie Rosenhain, Zuzanna~A Magnuska, Grace~G Yamoah, Wa’ Rawashdeh, Fabian
  Kiessling, Felix Gremse, et~al.,
\newblock ``A preclinical micro-computed tomography database including 3d whole
  body organ segmentations,''
\newblock {\em Scientific data}, vol. 5, no. 1, pp. 1--9, 2018.

\bibitem{ji2022amos}
Yuanfeng Ji, Haotian Bai, Jie Yang, Chongjian Ge, Ye~Zhu, Ruimao Zhang, Zhen
  Li, Lingyan Zhang, Wanling Ma, Xiang Wan, et~al.,
\newblock ``Amos: A large-scale abdominal multi-organ benchmark for versatile
  medical image segmentation,''
\newblock {\em arXiv preprint arXiv:2206.08023}, 2022.

\bibitem{luo2024rethinking}
Xiangde Luo, Zihan Li, Shaoting Zhang, Wenjun Liao, and Guotai Wang,
\newblock ``Rethinking abdominal organ segmentation (raos) in the clinical
  scenario: A robustness evaluation benchmark with challenging cases,''
\newblock 2024.

\bibitem{wasserthal2023totalsegmentator}
Jakob Wasserthal, Hanns-Christian Breit, Manfred~T Meyer, Maurice Pradella,
  Daniel Hinck, Alexander~W Sauter, Tobias Heye, Daniel~T Boll, Joshy Cyriac,
  Shan Yang, et~al.,
\newblock ``Totalsegmentator: robust segmentation of 104 anatomic structures in
  ct images,''
\newblock {\em Radiology: Artificial Intelligence}, vol. 5, no. 5, pp. e230024,
  2023.

\bibitem{hu2025towards}
Jiesi Hu, Yanwu Yang, Zhiyu Ye, Chenfei Ye, Hanyang Peng, Jianfeng Cao, and
  Ting Ma,
\newblock ``Towards robust in-context learning for medical image segmentation
  via data synthesis,''
\newblock {\em arXiv preprint arXiv:2509.19711}, 2025.

\bibitem{isensee2021nnu}
Fabian Isensee, Paul~F Jaeger, Simon~AA Kohl, Jens Petersen, and Klaus~H
  Maier-Hein,
\newblock ``nnu-net: a self-configuring method for deep learning-based
  biomedical image segmentation,''
\newblock {\em Nature methods}, vol. 18, no. 2, pp. 203--211, 2021.

\end{thebibliography}

\end{document}